# Debias-CLR: A Contrastive Learning Based Debiasing Method for Algorithmic Fairness in Healthcare Applications

Ankita Agarwal[1,2], Tanvi Banerjee[1,3], William L. Romine[1,3], Mia Cajita[4]
[1]*Department of Computer Science and Engineering, Wright State University*, Dayton, USA
[2]*Division of Biostatistics and Epidemiology, Cincinnati Children's Hospital Medical Center*, Cincinnati, USA
[3]*Kairos Research LLC*, Dayton, USA
[4]*College of Nursing, University of Illinois Chicago*, Chicago, USA
[1]{agarwal.15, tanvi.banerjee, william.romine}@wright.edu, [2]ankita.agarwal@cchmc.org, [3]{tanvi.banerjee, will}@kairosresearch.com,
[4]mcajit2@uic.edu

*Abstract*— Artificial intelligence based predictive models trained on the clinical notes of patients can be demographically biased, often influenced by the demographic distribution of the training data. This could lead to adverse healthcare disparities in predicting outcomes like length of stay of the patients. To avoid such possibilities, it is necessary to mitigate the demographic biases within these models so that the model predicts outcomes for individual patients in a fair manner. We proposed an implicit in-processing debiasing method to combat disparate treatment which occurs when the machine learning model predict different outcomes for individuals based on the sensitive attributes like gender, ethnicity, race, and likewise. For this purpose, we used clinical notes of heart failure patients and used diagnostic codes, procedure reports and physiological vitals of the patients. We used Clinical Bidirectional Encoder Representations from Transformers (Clinical BERT) to obtain feature embeddings within the diagnostic codes and procedure reports, and Long Short-Term Memory (LSTM) autoencoders to obtain feature embeddings within the physiological vitals. Then, we trained two separate deep learning contrastive learning frameworks, one for gender and the other for ethnicity to obtain debiased representations within those demographic traits. We called this debiasing framework as Debias-CLR. We leveraged clinical phenotypes of the patients identified in the diagnostic codes and procedure reports in the previous study to measure the fairness statistically. We found that Debias-CLR was able to reduce the Single-Category Word Embedding Association Test (SC-WEAT) effect size score when debiasing for gender from 0.8 to 0.3 and from 0.4 to 0.2 while using clinical phenotypes in the diagnostic codes and procedure reports respectively as targets. Similarly, after debiasing for ethnicity, the SC-WEAT effect size score reduced from 1 to 0.5 and from -1 to 0.3 in an opposite bias direction while using clinical phenotypes in the diagnostic codes and procedure reports respectively as targets. We further found that in order to obtain fair representations in the embedding space using Debias-CLR, the accuracy of the predictive models on downstream tasks like predicting length of stay of the patients did not get reduced as compared to using the un-debiased counterparts for training the predictive models. Hence, we conclude that our proposed approach, Debias-CLR is fair and representative in mitigating demographic biases and can reduce health disparities by making fair predictions for the underrepresented populations.

## I. INTRODUCTION

Electronic Health Records (EHRs) contain records of patients' health in a digital format. Earlier these records were typically used for billing purposes [1] but in recent years, these records have been leveraged to build artificial intelligence (AI) based predictive models to improve patients' outcomes [2]. However, these models can be demographically biased and can cause health disparities due to unfair decisions [3]. An EHR may include information on a patient's physiological vitals, administered medications, diagnostic codes, lab and procedure test results, as well as discharge summaries. Previous research has shown racial and ethnic disparities in diagnostic tests, therapeutic interventions [4], medications [5], and physiological vitals [6]. Likewise, gender bias has been observed in disease diagnosis [7], preventive therapies [8], and medications [9]. [10] found that machine learning models can detect a patient's self-reported race from clinical notes, even when explicit indicators of race were removed, while human experts were unable to do so. They also demonstrated that models trained on these race-related notes can continue to reinforce existing biases in clinical treatment decisions. Therefore, it is crucial to address demographic biases in these models and ensure fairness in AI-driven predictive systems.

We can attain fairness in these models by mitigating biases at different stages of their development by implementing pre-processing, in-processing and post-processing debiasing techniques. Pre-processing techniques involve modifying the data itself, such as through sampling, reweighting [11], or balancing data for demographically sensitive groups like gender, race, and ethnicity. In-processing techniques focus on building fair models by incorporating fairness considerations during model design, helping to reduce bias in feature embeddings [12] even when using biased data. Post-processing techniques adjust unfair model predictions by applying different thresholds for privileged and unprivileged groups. Among these, in-processing methods are particularly effective at reducing bias amplification caused by the algorithm during training [13]. Moreover, these methods can be applied to fine-tune representations from pre-trained debiased models, as the debiasing process is computationally

Research reported in this publication was supported in part funded by Department of Energy award number DE-EE0009097, National Center for Advancing Translational Sciences (NCATS), National Institutes of Health, through Grant Award Numbers UL1TR002003 and 1R01AT010413. Any content, opinions, findings, and conclusions or recommendations expressed in this material are solely the responsibility of the authors and do not necessarily reflect the official views of the DoE and NIH. The experimental procedures involving human subjects described in this paper were approved by the Institutional Review Board.

Corresponding author: Ankita Agarwal (agarwal.15@wright.edu)

intensive and requires significant effort. Therefore, developing a generalizable in-processing debiasing framework that can be applied across various domains, including healthcare, is essential.

In-processing debiasing methods to mitigate bias can be divided into explicit and implicit approaches. Explicit methods involve minimal adjustments to the objective functions, such as regularizing covariance relationships [14], absolute correlation regularization [15], or utilizing Wasserstein-1 distances [16]. Implicit mitigation methods typically involve deep learning techniques like adversarial learning [17], contrastive learning [18], and disentangled representation learning [19], which aim to debias representations so that predictions are free from discrimination against any underrepresented demographic group or individual. Both explicit and implicit methods can be further classified into two types: disparate impact and disparate treatment.

Disparate impact addresses fairness issues at the group level while disparate treatment addresses fairness issues at the individual level. Disparate impact addresses situations where the model predicts different outcomes for different demographic groups based on the attributes other than the sensitive attribute like gender, ethnicity, race etc., in the data set. In this case, fairness metrics that use statistical concepts to guarantee some form of parity to ensure fairness are measured [20]. Conversely, disparate treatment refers to cases where two otherwise similar data points, differing only in a sensitive attribute such as gender or ethnicity, receive different predictions from the model, suggesting that the model has unintentionally learned sensitive attribute information during training [21]. In this case, model fairness is assessed either through (a) fairness through awareness, where the difference in predictions between two individuals is constrained by the difference in their input features, ensuring that if two individuals are similar, their output differences are minimal, or through (b) counterfactual fairness, where counterfactual examples are created by reversing the sensitive attributes, so the model predicts the same outcome for both the actual individual and their counterfactual counterpart.

Deep learning methods are well-known for capturing complex non-linear patterns in data using multi-layer neural networks. Previous research has shown that implicit mitigation techniques effectively align bias reduction with fairness in these models [18]. Hence, in this work, we propose a neural disparate treatment based debiasing model, enabling the model to make fair predictions by considering individual-level similarities and differences among patients, regardless of sensitive attribute information, thereby improving the quality of healthcare for each patient.

To achieve algorithmic fairness in AI, there could be tradeoffs between the performance of a fair model on downstream tasks like predicting outcomes for a patient. Thus, it is important to evaluate the debiasing of the feature representations with respect to two important criteria [18]: (a) *fairness*, which measures the degree of bias in original and debiased feature embeddings, and (b) *representativeness*, which indicates the impact of debiased embeddings on downstream tasks as compared to the original embeddings. So, it is important to mitigate biases and improve fairness of these predictive models without impacting the robustness and performance of the model on downstream tasks.

Hence, we addressed the following research questions in this work:

1. Can we obtain debiased embeddings to mitigate demographic biases in the AI based predictive models and achieve algorithmic fairness in these models?
2. Can these debiased embeddings be applied to downstream tasks such as predicting length of stay of the hospitalized patients and compare in performance metrics to their un-debiased counterparts?

## II. RELATED WORK

Implicit debiasing methods focus on training deep learning networks to learn fair representations. These approaches include adversarial learning, disentangled representation learning, and contrastive learning. Adversarial and disentangled representation techniques are primarily aimed at reducing group-level disparities, making them focused on mitigating disparate impact [12]. In contrast, contrastive learning can address both group-level and individual-level differences, and may be either disparate impact driven or disparate treatment driven.

[22] developed an adversarial training framework to reduce algorithmic biases in clinical machine learning, targeting ethnicity and hospital location biases. They used data from the Oxford University Hospitals NHS Trust to predict COVID-19 cases during the first and second waves of the pandemic. The model used vital signs, blood tests, and other clinical features to screen COVID-19 patients while training two debiasing frameworks - one for ethnicity and another for hospital location. Their approach combined a predictor network for COVID-19 outcomes and an adversary network to mitigate bias, using multilayer perceptron models. The results showed improved fairness in terms of equalized odds while maintaining high clinical effectiveness, with a negative predictive value (NPV) greater than 0.98 for both sensitive features.

On the other hand, [23] applied a contrastive learning approach to achieve group-level fairness in the representation space, using disparate impact metrics to evaluate fairness. They trained a contrastive learning framework by pulling together instances with the same class label and pushing apart instances with the same sensitive attribute. The framework combined cross-entropy loss with two contrastive loss components and used the true positive rate gap (TPRG) as a fairness metric. They found a significant bias reduction in binary classification tasks like Twitter sentiment analysis (67.5%) and hate speech detection (25%), but less success with multiclass datasets like profession classification (TPRG increased by 0.1%) and image activity recognition (TPRG reduced by 11.5%). A limitation of their approach was its reliance on class labels and its ineffectiveness at addressing individual-level demographic biases.

Cheng et al. [18] introduced a neural debiasing method called FairFil, designed to address individual-level biases in pre-trained text encoders using a contrastive learning framework. Their method involved generating counterfactual examples for each training sentence by altering words to change bias direction without affecting the semantic meaning. The contrastive learning framework maximized cosine similarity between positive samples and minimized it between negative samples. They evaluated fairness using Sentence Encoder Association Test (SEAT) and found a 30% reduction in bias for pre-trained BERT embeddings and a 10% reduction in BERT sentiment classification tasks. However, their approach was limited to sentence embeddings, lacked applicability to datasets which include both text and numerical features, and was not generalizable to domains like healthcare where there are limited gender related words in the sentences as most of the data are related to clinical entities.

Thus, there are gaps in literature which we have proposed to address in this work by developing a generalized deep learning framework to mitigate implicit biases in machine learning models while accounting for individual level disparities. Hence, we propose to develop a novel self-supervised contrastive learning framework to account for individual level similarities and differences while learning fair representations. The rationale is that such a model can attain algorithmic fairness and representativeness by preserving the semantics for various downstream tasks like predicting length of stay while being demographically fair and trustworthy. Additionally, such a framework can be utilized to attain fairness through counterfactual examples where sensitive attribute related words like he, she, him, her are absent and the data is not limited to text.

## III. METHODS

### A. Dataset Collection

We accessed the EMR of 1,200 patients admitted to UI Health for heart failure between August 2016 and August 2021, all of whom were 18 years or older at the time of their first hospitalization, through the UI Chicago Center for Clinical and Translational Science (CCTS) Biomedical Informatics Core. During each hospitalization, data on diagnostic codes (both primary and secondary) following the International Classification of Diseases, Tenth Revision, Clinical Modification (ICD-10 CM) system, procedures performed, and physiological vitals were recorded. To integrate the diagnostic codes and procedure reports for each patient's hospitalization, we employed the same method as described in our previous study [24].

To summarize the method used in our previous study, we first combined the primary and secondary diagnosis codes of a patient during each hospitalization and converted these to text using the Clinical Classifications Software Refined dataset created by the Agency for Healthcare Research and Quality[1] which aggregates more than 70,000 ICD-10-CM diagnosis codes into over 530 clinically meaningful categories. Thereafter, we removed the word '*Heart Failure*' as this word would not contribute to our further analysis. We extracted the impression or conclusion attribute from the procedure reports of the patient. If the notes did not contain either of these attributes, we extracted the findings attribute from these notes for the analysis and discarded the notes which did not contain either the impression/conclusion or finding attribute. During each hospitalization, a patient's information about the diagnostic codes remained the same, whereas there could be more than one record for the procedure reports. So, we concatenated all the procedure reports of the patient during each hospitalization.

Thereafter, we compiled information about the physiological vitals of the patients which were collected during each hospitalization. These vitals include systolic blood pressure (in mmHg), diastolic blood pressure (in mmHg), SpO2 (in percentage), temperature (in Fahrenheit), pulse rate (in beats per minute), and respiratory rate (breaths per minute). Earlier studies have shown that these six physiological vitals can help to get insight into a patient's condition and can be used for predictive modeling [25].

### B. Dataset Preparation

We used the clinical BERT model [26] to obtain embeddings for the diagnostic codes and procedure reports to learn contextual meaning between their words. As all the procedure reports for each patient during a particular hospitalization were concatenated, clinical BERT could learn relationships between the terms from the day of admission to discharge. For this purpose, we used Hugging Face Clinical BERT implementation[2] to get 768 dimensional embeddings each for diagnostic codes and procedure reports.

There were multiple records of each measure during each visit/hospitalization of the patient as the vitals were typically measured after some interval of time to monitor the condition of the patient. Long Short-Term Memory (LSTM) neural networks have been known to model temporal sequences [27]. Hence, we used Long Short-Term Memory (LSTM) autoencoders to obtain embeddings for each physiological vital. The LSTM autoencoders learn a compressed representation of a temporal sequence data by using the encoder-decoder LSTM architecture. We used Python Keras LSTM architecture for this purpose[3]. We created the LSTM model with two LSTM layers of 100 neurons and RELU activation function to obtain 100 dimensional embeddings for each physiological vital.

Once we had obtained the embeddings for diagnostic codes, procedure reports and physiological vitals using the methods, we concatenated these embeddings together corresponding to each patient hospitalization. So, the total feature size had an embedding dimension of 2,136 for each patient hospitalization record. Further, we concatenated the gender, ethnicity, and length of stay information to each record. The length of stay was also calculated using the criteria as explained in our previous study [24]. Length of stay means the number of days between the hospital admission and discharge date of the patient during a particular stay. So, based on the number of days, we categorized these into five categories: very short-term stay (0-1 day or 24 hours hospitalization), short-term stay (2-7 days), medium-term stay (8-14 days), long-term stay (15-21 days) or very long-term stay (more than 21 days).

For each record, we also appended two 12-dimensional one hot encoding vectors corresponding to the 12 phenotypes identified in the diagnostic codes and procedure reports such as 'Chronic Heart Disease', 'Myocardial Ischemia', 'Hypertrophic Cardiomyopathy', and so on as determined in our previous study [24]. Hence, for example, if a sample belonged to the 'Chronic Heart Disease' phenotype based on its diagnostic code and 'Myocardial Ischemia' based on its procedure reports, the value corresponding to these two phenotypes were 1, and for the rest phenotypes, the value was 0. Final, we were left with a total of 2,429 hospitalization records of 1,200 patients. The number of records were 261, 1319, 542, 164, and 143 for very short-term stay, short-term stay, medium-term stay, long-term stay, and long-term stay respectively. There were a total of 1,269 records for female and 1,160 records for male patients. Similarly, there were 510 records for Hispanics and 1,919 records for non-Hispanic patients.

---

[1] https://www.hcup-us.ahrq.gov/toolssoftware/ccsr/ccs_refined.jsp
[2] https://huggingface.co/medicalai/ClinicalBERT
[3] https://keras.io/api/layers/recurrent_layers/lstm/

*C. Contrastive Learning Framework*

Contrastive learning (CL) is a self-supervised approach for learning representations by distinguishing between positive and negative samples. The underlying assumption is that similar samples should be positioned closer together in the representation space, while dissimilar samples should be farther apart. The objective of contrastive learning is to increase the cosine similarity between a sample and its positive counterpart, while decreasing the cosine similarity with negative samples, as defined by the loss function in Equation 1:

$$l_{i,j} = -\log \frac{\exp(\text{sim}(z_i, z_j)/\tau}{\sum_{k=1}^{2N}[k \neq i]\exp(\text{sim}(z_i, z_k)/\tau} \quad (1)$$

where, $z_i$ and $z_j$ are the positive samples and $z_i$ and $z_k$ are the negative samples.

We utilized this framework and executed it with a batch size of 1024 for 100 epochs using the LARS optimizer (Layer-wise Adaptive Rate Scaling) [28] to achieve fair representations and obtain debiased feature embeddings within the latent space. We refer to this framework as Debias-CLR.

*D. Feature Selection for Sensitive Attribute*

In this study, we trained two separate contrastive learning frameworks to address two sensitive attributes: gender and ethnicity. The gender attribute had two classes - female and male, while ethnicity was divided into Hispanic and non-Hispanic classes. To train the contrastive learning models, we first identified the top 50% of features which were best at predicting the sensitive attribute and called these features as 'sensitive features.'

Among the 2,429 samples, there were 1,269 female and 1,160 male records. For ethnicity, there were 510 Hispanic and 1,919 non-Hispanic records. The gender distribution was relatively balanced, while the ethnicity distribution was notably imbalanced. We addressed this imbalance in both the sensitive attributes using the Synthetic Minority Oversampling Technique (SMOTE) [29]. To select the sensitive features, we utilized Mutual Information (MI), which measures the dependency between two variables, with higher values indicating stronger dependency. MI was employed to identify features most predictive of gender when aiming to obtain debiased representations with respect to gender, or most predictive of ethnicity when the goal was to obtain debiased representations with respect to ethnicity.

After selecting the sensitive features using the MI score, we used five supervised machine learning algorithms to predict the sensitive attribute based on these features, using an 80:20 train-test split ratio. The machine learning (ML) models included k-Nearest Neighbors (kNN), logistic regression (LR), support vector machines (SVM), multilayer perceptron (MLP), and random forest (RF). The sensitive features predicted gender with an accuracy of 70.2% using LR and ethnicity with an accuracy of 86.3% using RF. After identifying the sensitive features for both gender and ethnicity, we used these to generate positive samples for each instance to train the contrastive learning framework, as described in the next section.

*E. Generative counterfactual Examples*

We propose a novel method to generate counterfactual examples which would constitute positive and negative samples in the contrastive learning framework for debiasing. This method is generalizable to different domains including healthcare and is not limited to text data for generating counterfactual examples.

Once, we had identified the sensitive features for the sensitive attribute either gender or ethnicity, we generated counterfactual examples and identified positive and negative samples for each sample to train two separate contrastive learning frameworks, one to obtain debiased representation to mitigate gender-related bias and another one to obtain debiased representation to mitigate ethnicity-related bias, using Algorithm 1. In Algorithm 1, as mentioned earlier, sensitive attribute gender has two classes, class 1 as female and class 2 as male. Similarly, sensitive attribute ethnicity has two classes, class 1 for Hispanic and class 2 as non-Hispanic.

| **Algorithm 1:** Algorithm to generate positive and negative samples through counterfactual examples | |
|---|---|
| | **Input:** Feature embedding vector of all features (A), sensitive features (X), sensitive attribute (S) for each sample |
| | **Output:** Positive and negative samples corresponding to each sample |
| | **count** ← Number of samples |
| 1 | Initialize i to 0 |
| 2 | For the first class of the sensitive attribute, find the average value of each sensitive feature and make a vector Xf = {Xf$_1$, Xf$_2$, Xf$_3$, ……………., Xf$_n$}, where n is the number of sensitive features |
| 3 | For the second class of the sensitive attribute, find the average value of each sensitive feature and make a vector Xs = {Xs$_1$, Xs$_2$, Xs$_3$, ……………., Xs$_n$}, where n is the number of sensitive features |
| 4 | **while** (i < count) do |
| 5 |     if S[i] = = 'class 1' //if the sample belonged to the first class of sensitive attribute |
| 6 |         Values of all features other than the sensitive features remain same |
| 7 |         Replace the values for sensitive features with the values in vector Xs<br>This will form a positive sample |
| 8 |     else if S[i] = = 'class 2' //if the sample belonged to the second class of sensitive attribute |
| 9 |         Values of all features other than the sensitive features remain same |
| 10 |         Replace the values for sensitive features with the values in vector Xf<br>This will form a positive sample |
| 11 |     All other samples will form the negative samples |
| 12 |     Increment i by 1 |
| | **end** |

*F. Fairness Metric – SC-WEAT*

The Word Embedding Association Test (WEAT) metric [30] is primarily utilized in natural language processing to evaluate the association between two traditional and contextual semantic representations of words, known as word embeddings, in relation to two targets and two attributes. For instance, in a straightforward social context, the attributes might consist of words frequently associated with each gender, while the targets could be words commonly linked to each gender in a stereotypical scenario. The WEAT metric operates under a null hypothesis that posits no difference between the two sets of target words regarding their relative similarity to the two sets of attribute words. Thus, it measures the differential association between a specific pair of targets and an attribute.

In our previous work, we used the clinical notes of patients admitted for heart failure to identify the themes within these notes to infer severity of heart failure, cardiovascular comorbidities associated with patients admitted for heart failure [31] and to predict length of stay of the patients [24]. Hence, in this work, we used the clinical phenotypes identified in the diagnostic codes and procedure reports in the form of themes as targets [24]. As the clinical phenotypes remain the same for both classes of the sensitive attribute, in our study, there was only one class of target instead of two. So, for this purpose, we used insights from Single-Category WEAT or SC-WEAT [32] to calculate the effect size and association between target and attributes. The effect size for SC-WEAT is calculated as shown in Equation 2 and Equation 3.

$$s(T1,T2,A1,A2) = \sum_{x \in T1} s(x,A1,A2) - \sum_{y \in T2} s(y,A1,A2) \quad (2)$$

where,

$$s(w,A1,A2) = \text{mean}_{(a \in A1)} \cos(w,a) - \text{mean}_{(b \in A2)} \cos(w,b) \quad (3)$$

The SC-WEAT effect size can range from -2 to 2 [32]. A negative value indicates a stronger association of the word with attribute A1, while a positive value indicates a stronger association with attribute A2. Ideally, for an unbiased fair model, the SC-WEAT effect size should be close to 0. After training two contrastive learning frameworks - one for gender and the other for ethnicity, we computed the fairness metric, SC-WEAT, using Algorithm 2 to measure the reduction in bias related to gender and ethnicity within the contrastive learning framework.

### G. Feature Regularization

To test the representation and robustness of the debiased embeddings, we added feature regularization in the input space using cutout strategy. Cutout is a regularization technique in which some sections of the input data are removed so that the model learns representations by considering all the features during training rather than just some key features. This helps the model to learn robust and contextualized representations and prevents overfitting. We masked a random 20% of the sensitive features for each sample, converting their values as zero. Additionally, we tested this approach on different chunks of the data to use 20%, 40%, 60%, and 80% of the data as training data. We trained the Debias-CLR model on each of these chunks and calculated the SC-WEAT effect size scores. We treated Debias-CLR as a baseline approach and named Debias-CLR with cutout regularization model as Debias-CLR-R.

### H. Performance Metrics

In addition to the traditional performance metrics like accuracy we also used Matthews Correlation Coefficient (MCC) and Cohen's Kappa (K) to test the representativeness of the fair model. So, we used binary categories for length of stay and merged the samples belonging to classes 1 and 2 into one category and the samples belonging to classes 3, 4, and 5 into another category. As there was an imbalance in the outcomes like length of stay, we used SMOTE to balance the categories and compared the performance of the biased model in predicting length of stay with Debias-CLR and Debias-CLR-R using five ML algorithms which were k-Nearest Neighbors (kNN), logistic regression (LR), support vector machines (SVM), multilayer perceptron (MLP), and random forest (RF).

MCC is an important metric in context of contrastive learning as it helps to evaluate the effectiveness of the debiased representations in predicting samples from both the classes equally well [33]. It considers all four blocks of the confusion matrix, and its value lies between -1 and 1. Similarly, Cohen's Kappa (K) takes agreement by chance into consideration [34]. Earlier it was proposed to test interrater reliability but it has been introduced as a performance metric and its value lies between 0 to 1. The value of Cohen's Kappa value as 0 indicates no agreement and 0.01 - 0.20 as none to slight, 0.21 - 0.40 as fair, 0.41 - 0.60 as moderate, 0.61 - 0.80 as substantial, and 0.81 - 1.00 as almost perfect agreement. For example, in this study, a value of 0.7 would mean that the model predicted samples for both classes with a substantial equality.

| Algorithm 2: Algorithm to Calculate SC-WEAT effect size |
|---|
| **Input: Two datasets** |
| First dataset (D1) with 2136-dimensional Feature embeddings of the training set before Debias-CLR ($X_B$), sensitive attribute (gender of a sample if debiasing for gender else ethnicity of a sample if debiasing for ethnicity), two 12-dimensional one hot encoding vectors corresponding to the 12 phenotypes identified in the diagnostic codes and procedure reports |
| Second dataset (D2) with 2136-dimensional Feature embeddings of the training set after Debias-CLR ($X_A$), gender, two 12-dimensional one hot encoding vectors corresponding to the 12 phenotypes identified in the diagnostic codes and procedure reports |
| **Output:** SC-WEAT effect size before and after Debias-CLR |

| | |
|---|---|
| 1 | Using D1, separate samples of two classes of the sensitive attribute |
| 2 | A1 ← 2136-dimensional Feature embeddings corresponding to first class |
| 3 | A2 ← 2136-dimensional Feature embeddings corresponding to second class |
| 4 | if (target == 'Diagnostic codes') |
| 5 |    T ← 12-dimensional one hot encoding vectors corresponding to the phenotypes in the diagnostic codes |
| 6 |    $\text{mean}_1 = \text{mean}_{a \in A1 \text{ and } w \in T} \cos(w,a)$ |
| 7 |    $\text{mean}_2 = \text{mean}_{a \in A2 \text{ and } w \in T} \cos(w,a)$ |
| 8 |    $\text{stddev} = \text{stddev}_{x \in A1 \cup A2 \text{ and } w \in T} \cos(w,x)$ |
| 9 |    SC-WEAT effect size = $(\text{mean}_1 - \text{mean}_2) / \text{stddev}$ |
| | Repeat Steps 6 to 10 if (target == 'Procedure reports') and T ← 12-dimensional one hot encoding vectors corresponding to the phenotypes in the procedure reports |
| | Repeat Steps 1 to 9 using D2 |

The overall methodology of the proposed work is shown in Fig. 1.

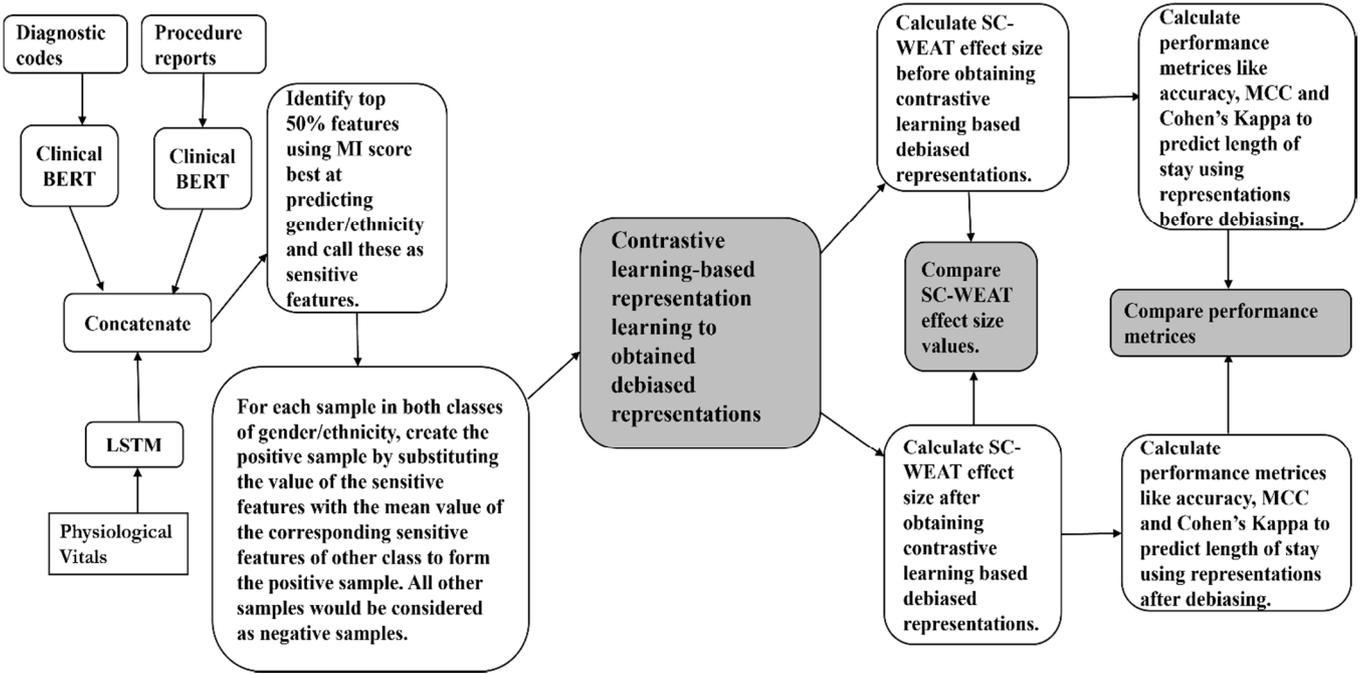

Fig. 1. Diagrammatic representation of the proposed methodology to obtain debiased representations and to compare the fairness and representativeness of these debiased representations with their un-debiased counterparts

## IV. RESULTS

### A. Feature Selection

The results of the five supervised machine learning algorithms to predict gender and ethnicity in terms of accuracy are shown in Table I.

We found that LR performed the best at predicting gender using the sensitive features with an accuracy of 70.2% while for ethnicity, RF performed the best at predicting ethnicity using the sensitive features with an accuracy of 86.3%. For both the sensitive attributes, selecting top 50% of the sensitive features improved an accuracy of predicting the respective classes, indicating a relationship between the features' representation and sensitive attribute, which a machine learning algorithm can learn while training predictive models. A higher accuracy of 86.3% in predicting ethnicity indicates a higher relationship between the features and ethnicity classes. Thus, it is necessary to remove bias in the predictive models by mitigating this relationship to attain algorithmic fairness in the representations.

### B. SC-WEAT Effect Size

The results of calculating SC-WEAT effect size are shown in Table II. It was found that as the amount of training data increased, the value of SC-WEAT effect size got reduced which indicates that to build a fair model using contrastive learning framework, more examples for positive and negative samples are required. With just 20% of the training data, the SC-WEAT effect size increased but gradually it decreased as more training samples were added.

### C. Length of Stay Prediction

The results in Tables III and IV show that the random forest algorithm achieved the best performance on biased representations, using 80% of the training data, suggesting a non-linear relationship between the features. Notably, after applying debiasing, logistic regression outperformed other models in predicting length of stay, indicating that the proposed framework introduced linearity between the features through specific transformations. The debiasing frameworks for gender and ethnicity did not compromise the accuracy of length of stay predictions, and feature regularization further enhanced the predictive performance for length of stay.

## V. DISCUSSION

The proposed framework Debias-CLR is an in-processing method to address disparate treatment at the individual level by implementing a self-supervised contrastive learning model to obtain debiased embeddings in the clinical data to mitigate biases and improve fairness of the predictive models without impacting the performance of the model on downstream tasks. We tested the debiased embeddings obtained to achieve algorithmic fairness for predicting length of stay of hospitalized heart failure patients with no trade off on accuracy. This supported our initial hypothesis that the embedding space was vast enough to generate debiased representations without affecting performance in potential downstream tasks. Our method of generating counterfactual examples was not limited to text and sentences as seen in previous CL based debiasing models [18] but is generalizable to additional modalities of data like numerical features. We were also able to generate positive and negative samples to train the CL framework, going beyond explicit sensitive attribute specific words like 'he', 'she' (instances of gender disparity), or words like 'Hispanic', 'non-Hispanic' (instances of ethnic disparity). This is significant since, in order to generate counterfactual examples, these sensitive attribute specific words might not be present and limited to text based data in healthcare and other domains. Moreover, if we are using multimodal data like vitals, lab results, medical images, these sensitive attribute related features within these data would be altogether absent, creating a need for more implicit and nuanced bias detection mechanisms such as Debias-CLR.

TABLE I. PREDICTION **ACCURACY** OF THE SENSITIVE ATTRIBUTE USING ALL FEATURES AND SENSITIVE FEATURES

| Sensitive attribute (target) | Features | kNN | LR | SVM | MLP | RF |
|---|---|---|---|---|---|---|
| Gender | All | 0.546 | 0.650 | 0.591 | 0.566 | 0.625 |
| | Sensitive features | 0.591 | **0.702** | 0.609 | 0.607 | 0.603 |
| Ethnicity | All | 0.668 | 0.767 | 0.725 | 0.832 | 0.841 |
| | Sensitive features | 0.711 | 0.792 | 0.698 | 0.854 | **0.863** |

TABLE II. SC-WEAT EFFECT SIZES BETWEEN ATTRIBUTES AND TARGET WITH DIFFERENT % OF TRAINING DATA

| Sensitive Attribute | % of Training Data | Diagnostic codes phenotypes | | | Procedure Reports phenotypes | | |
|---|---|---|---|---|---|---|---|
| | | Before | Debias-CLR | Debias-CLR-R | Before | Debias-CLR | Debias-CLR-R |
| Gender | 20% | 1.2 | 1.4 | 1.2 | 0.5 | 0.5 | 0.5 |
| | 40% | 1.2 | 0.8 | 0.7 | 0.8 | 0.4 | 0.4 |
| | 60% | 1.2 | 0.4 | 0.3 | 0.5 | 0.3 | 0.2 |
| | **80%** | **0.8** | **0.3** | **0.2** | **0.4** | **0.2** | **0.2** |
| Ethnicity | 20% | 1.0 | 1.4 | 1.2 | -0.4 | -0.6 | -0.5 |
| | 40% | 0.6 | 0.7 | 0.5 | 0.6 | 0.4 | 0.3 |
| | 60% | 1.2 | 0.6 | 0.5 | 1 | -0.3 | 0.3 |
| | **80%** | **1** | **0.5** | **0.3** | **-1** | **0.3** | **0.2** |

Previous studies have used metrics like WEAT (in case of two targets) [18], or SC-WEAT (in case of single target) [32] as the fairness metric to calculate the effect size of bias present in the predictive model. Usually, the attribute classes in SC-WEAT contain different features for two classes for sensitive attribute like gender or ethnicity, but in this work, we have modified SC-WEAT to remove this constraint such that the features in both the attribute classes could be the same but their feature embedding vectors would be different. Removing such a constraint is important in chronic health conditions such as heart failure. For example, the role of an important feature like heart rate could be an important attribute to measure both for male and female in case of target such as Arrhythmia, but its values is different for each gender [35]. Hence, the value of the mean of association of the heart rate with Arryhthmia would be different for male and female patients introducing bias within the dataset. However, Debias-CLR can reduce this difference and improve fairness of the predictive models by obtaining debiased feature representations, so that these models do not predict different outcomes for male and female patients having same un-debiased feature embeddings and only differing in the sensitive attribute like gender. Hence, the proposed approach can address both bias and fairness, an issue which most debiased models fail to address.

Earlier studies have treated SC-WEAT as analogous to Cohen's d [32]. Our findings show that Debias-CLR effectively reduced SC-WEAT effect size values for associations between features and clinical phenotypes across both classes of sensitive attributes, such as gender and ethnicity, bringing them closer to 0. For gender, the mean association between feature embeddings and diagnostic codes or procedure reports for the female class was higher than for the male class, revealing a significant disparity and bias toward females. For ethnicity, the mean association between feature embeddings and diagnostic codes was higher for the Hispanic class compared to the non-Hispanic class, while the reverse was true for procedure reports. This aligns with previous research showing that Hispanic patients tend to receive fewer procedures than non-Hispanic patients [36], suggesting that the representations had unintentionally learned implicit biases. After applying Debias-CLR, we successfully reduced these disparities, achieving both algorithmic fairness and bias mitigation. This kind of a study evaluating the impact of debiasing on effect size has not been previously conducted

TABLE III. COMPARISON OF PERFORMANCE OF BIASED REPRESENTATIONS WITH DEBIAS-CLR AND DEBIAS-CLR-R TO MITIGATE GENDER BIAS FOR PREDICTING LENGTH OF STAY

| Model | Embeddings | A | MCC | K |
|---|---|---|---|---|
| kNN | Raw Embeddings | 0.632 | 0.376 | 0.376 |
| | Debias-CLR | 0.612 | 0.369 | 0.369 |
| | Debias-CLR-R | 0.610 | 0.359 | 0.359 |
| **LR** | **Raw Embeddings** | **0.743** | **0.487** | **0.487** |
| | **Debias-CLR** | **0.765** | **0.502** | **0.502** |
| | **Debias-CLR-R** | **0.780** | **0.511** | **0.511** |
| SVM | Raw Embeddings | 0.724 | 0.475 | 0.455 |
| | Debias-CLR | 0.736 | 0.479 | 0.479 |
| | Debias-CLR-R | 0.741 | 0.482 | 0.482 |
| MLP | Raw Embeddings | 0.711 | 0.395 | 0.395 |
| | Debias-CLR | 0.726 | 0.409 | 0.409 |
| | Debias-CLR-R | 0.725 | 0.398 | 0.398 |
| RF | Raw Embeddings | 0.760 | 0.520 | 0.520 |
| | Debias-CLR | 0.737 | 0.504 | 0.504 |
| | Debias-CLR-R | 0.722 | 0.494 | 0.494 |

Note: A – Accuracy, MCC – Mathews Correlation Coefficient, K- Cohen's Kappa

TABLE IV. COMPARISON OF PERFORMANCE OF BIASED REPRESENTATIONS WITH DEBIAS-CLR AND DEBIAS-CLR-R TO MITIGATE ETHNICITY BIAS FOR PREDICTING LENGTH OF STAY

| Model | Embeddings | A | MCC | K |
|---|---|---|---|---|
| kNN | Raw Embeddings | 0.723 | 0.398 | 0.398 |
| | Debias-CLR | 0.712 | 0.387 | 0.387 |
| | Debias-CLR-R | 0.711 | 0.386 | 0.386 |
| **LR** | **Raw Embeddings** | **0.817** | **0.536** | **0.536** |
| | **Debias-CLR** | **0.842** | **0.552** | **0.552** |
| | **Debias-CLR-R** | **0.858** | **0.562** | **0.562** |
| SVM | Raw Embeddings | 0.798 | 0.516 | 0.516 |
| | Debias-CLR | 0.801 | 0.518 | 0.518 |
| | Debias-CLR-R | 0.803 | 0.518 | 0.518 |
| MLP | Raw Embeddings | 0.784 | 0.498 | 0.498 |
| | Debias-CLR | 0.798 | 0.501 | 0.501 |
| | Debias-CLR-R | 0.801 | 0.506 | 0.506 |
| RF | Raw Embeddings | 0.836 | 0.572 | 0572 |
| | Debias-CLR | 0.811 | 0.555 | 0.555 |
| | Debias-CLR-R | 0.794 | 0.543 | 0.543 |

Note: A – Accuracy, MCC – Mathews Correlation Coefficient, K- Cohen's Kappa

by researchers, but it is crucial for assessing and quantifying how much a specific clinical phenotype may be associated with a particular gender or ethnicity. This, in turn, could influence treatment decisions if inherent bias exists in the predictive model, which is especially important in healthcare applications.

We also refuted the argument that debiasing a predictive model could adversely impact downstream tasks, i.e. create a fairness vs accuracy trade-off [18]. Finally, as performance metrics like accuracy, precision, and recall might not be sufficient to study the ability of the algorithm in predicting samples from both classes with a comparable equality, we tested the performance of Debias-CLR using MCC and Cohen's Kappa.

Debias-CLR was able to improve algorithmic fairness as more training samples were introduced, and at the same time, achieved linearity within the relationship between features, thereby reducing the computational time complexity of the model. Prior to debiasing, the random forest classifier performed the best in predicting length of stay, but logistic regression performed the best for downstream task after debiasing, which indicates that Debias-CLR was able to reduce the computational time complexity for predicting outcomes.

An improvement of about 2.5% in predicting outcomes using Debias-CLR framework for ethnicity indicates that before debiasing, there were complex relationships within the representations of features which were tightly bound to the sensitive attribute like ethnicity. This supports the existing literature [37] that there is a higher disparity among individuals from different ethnicities seeking health care during their hospitalization. Building a disparate treatment framework for individuals based on their ethnicity like Debias-CLR plays an important role in improving predictive ability for machine learning models. Adding feature regularization using cutout strategy reduced the SC-WEAT effect size scores further, indicating fairer model. Feature regularization also improved the MCC and Cohen's Kappa score for predicting length of stay. This indicated that logistic classifier trained on debiased representations using our proposed framework obtained symmetry in the confusion matrix and thus predicted the outcomes of both classes with an equal accuracy.

This study has its own limitations. In this study, we used the dataset from a single hospital where there were no missing records for the gender and ethnicity but there could be instances where this data is missing, not reported or could not be disclosed. It is important to test the generalizability and robustness of the proposed framework on diverse datasets and patient papulation groups from different hospitals outside the Chicago area, and medical conditions other than heart failure. Additionally, as contrastive learning framework approaches may face challenges with data that includes biased correlations [38], hierarchical or temporal relationships among the features, it is necessary to perform error analysis to explore such scenarios to test the performance and reliability of Debias-CLR.

Integrating Debias-CLR into a real-world healthcare system would also require careful planning due to sensitive data, regulatory compliance guidelines like the Health Insurance Portability and Accountability Act (HIPAA) which limits access to certain datasets. It would also be necessary to conduct rigorous validation and scalability of Debias-CLR in simulated clinical environments using real-world high-dimensional data. The debiased model should be integrated with EHRs to streamline data flow and minimize disruptions to the clinical workflows and decision-making processes which would require technical compatibility with the existing systems. Hence, it would require collaboration with the clinicians and the deployment team. Additionally, periodic retraining and bias audits of the model are necessary to be performed to detect shifts in the bias over time as the data changes to prevent performance degradation due to emerging biases.

In the future, we might need to implement time-series based data augmentation techniques [39] to generate counterfactual examples for debiasing since previous studies have shown that healthcare forecasting varies over time based on demographic features [40]. Earlier studies have also demonstrated that as the themes identified within the dataset could vary over time [41], [42], [43], the clinical phenotypes identified in the form of themes which were used to measure SC-WEAT effect size could also vary over time. This could also adversely affect the performance of the predictive algorithms on different downstream tasks. Hence, in future we need to calculate the SC-WEAT effect size scores over time.

## VI. CONCLUSION

There could be inherent bias in the feature representations of the model in the latent space which needs to be addressed in order to improve algorithmic fairness in the predictive models especially in healthcare where an unfair model can lead to health disparities in different demographic sectors. We have proposed a novel contrastive learning-based framework known as Debias-CLR to address this disparity and address fairness at the individual level using counterfactual examples. We evaluated the reduction in the bias by modifying SC-WEAT effect size to calculate the association between the feature embeddings and clinical phenotypes and found a reduction in the SC-WEAT effect sizes for both gender and ethnicity indicating fair representations. Our proposed framework for debiasing, Debias-CLR and Debias-CLR-R did not cause a reduction in the accuracy of predicting downstream tasks like length of stay; in fact, the Debias-CLR-R outperformed the earlier model by 4.1% using the accuracy metric. Furthermore, it reduced the computational time complexity for predictive tasks in addition to obtaining algorithmic fairness in AI based predictive models in healthcare domain. Our study makes a strong case for building debiased algorithms in predictive tasks in the healthcare domain, advocating healthcare for all and making sure that no underrepresented groups receive inappropriate care in the future.